\documentclass{article}
\usepackage{ijcai21}

\usepackage{times}
\usepackage{soul}
\usepackage{url}
\usepackage[hidelinks]{hyperref}
\usepackage[utf8]{inputenc}
\usepackage[small]{caption}
\usepackage{graphicx}
\usepackage{amsmath}
\usepackage{amsthm}
\usepackage{booktabs}
\usepackage{algorithm}
\usepackage{algorithmic}
\title{Learning Embeddings from Knowledge Graphs With Numeric Edge Attributes}

\author{
Sumit Pai
\and
Luca Costabello
\affiliations
Accenture Labs\\
\emails
\{sumit.pai, luca.costabello\}@accenture.com
}

\usepackage{hyperref}
\usepackage{url}
\usepackage[english]{babel}
\usepackage{blindtext}
\usepackage{multirow}
\usepackage{booktabs}
\usepackage{soul}
\usepackage{mathtools}
\usepackage{amsmath}
\usepackage{amssymb}
\usepackage{todonotes} 
\usepackage{xcolor}
\usepackage{color}
\usepackage{tabularx}
\usepackage{makecell}
\usepackage{caption}
\usepackage{subcaption}

\usepackage[switch]{lineno}

\usepackage{siunitx}
\usepackage{amsmath,amsfonts,bm}

\def\eqref#1{equation~\ref{#1}}

\def\1{\bm{1}}

\DeclareMathAlphabet{\mathsfit}{\encodingdefault}{\sfdefault}{m}{sl}
\SetMathAlphabet{\mathsfit}{bold}{\encodingdefault}{\sfdefault}{bx}{n}

\newcommand{\R}{\mathbb{R}}

\newcommand{\Real}{\ensuremath{\mathbb{R}}}

\begin{document}

\maketitle

\begin{abstract}
Numeric values associated to edges of a knowledge graph have been used to represent uncertainty, edge importance, and even out-of-band knowledge in a growing number of scenarios, ranging from genetic data to social networks. Nevertheless, traditional knowledge graph embedding models are not designed to capture such information, to the detriment of predictive power.
We propose a novel method that injects numeric edge attributes into the scoring layer of a traditional knowledge graph embedding architecture. Experiments with publicly available numeric-enriched knowledge graphs show that our method outperforms traditional numeric-unaware baselines as well as the recent UKGE model.
\end{abstract}

\blindmathtrue

\section{Introduction}\label{sec:intro}

Knowledge graphs are graph-based knowledge bases whose facts are modeled as labeled, directed edges between entities. Whether it is a social network, a bioinformatics dataset, or retail purchase data, modelling knowledge as a graph lets organizations capture patterns that would otherwise be overlooked. Research led to broad-scope graphs such as DBpedia~\cite{auer2007dbpedia}, WordNet, and YAGO~\cite{suchanek2007yago}. Countless domain-specific knowledge graphs have also been published on the web, giving birth to the so-called Web of Data~\cite{bizer2011linked}.

Knowledge graph embeddings (KGE) are a family of graph representation learning methods that learn vector representations of nodes and edges of a knowledge graph. They are applied to graph completion, knowledge discovery, entity resolution, and link-based clustering, just to cite a few~\cite{nickel2016review}. 

\begin{figure}
\centering
\includegraphics[width=.7\columnwidth]{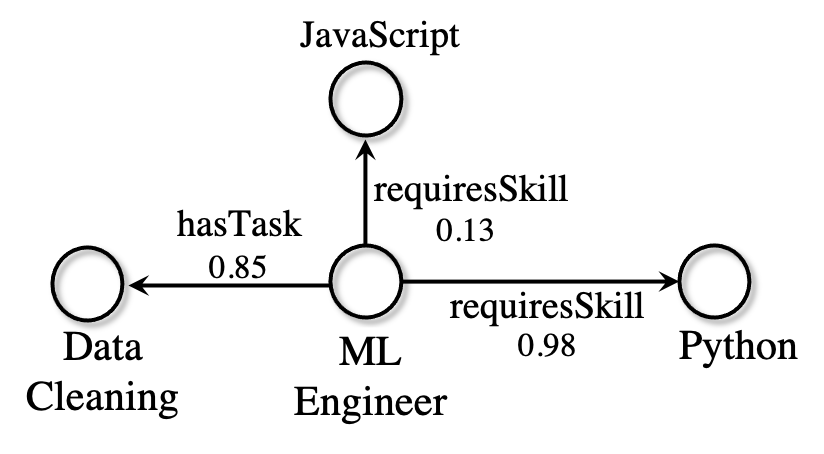}
\caption{A Knowledge graph with numeric attributes associated to triples.}
\label{fig:kg}
\end{figure}

In \textit{multimodal} knowledge graphs, entity nodes are associated to attributes from different data modalities, such as numbers, text, images. 
In this paper we deal with a specific flavor of multimodal knowledge graphs, that is graphs with numeric-enriched edges, either at predicate-level (i.e. a number assigned to each predicate \textit{type}), or specific to each triple (this latter case is shown in Figure~\ref{fig:kg}).
Numeric-enriched triples have a prominent role in a growing number of applicative scenarios, from protein networks to workforce knowledge bases. Such numeric values may represent edge uncertainty (i.e. uncertain knowledge graphs) as in ConceptNet~\cite{speer2017conceptnet}; trust in the automated or semi-automated data extraction workflow used to build the graph~\cite{mitchell2018never}; out-of-band knowledge from wet lab experiments~\cite{PPI}; edge importance or link strength.

A number of works in knowledge graph representation learning literature support multimodal information and leverage numeric values associated to node entities to generate better embeddings for enhanced link prediction~\cite{garcia2017kblrn,kristiadi2019incorporating,wu2018knowledge}.
Nevertheless, such models are not designed to learn from numeric values associated to \textit{edges} of a knowledge graph. With the notable exception of \cite{chen2019embedding}, that however is designed for uncertain graphs only, supporting numeric is to date still an under-researched direction.

In this work, we focus on the task of predicting probability estimates of missing links in knowledge graphs with numeric-enhanced triples. We claim that a model must take into account such numeric literals. Regardless of their semantics, we operate under the assumption that such numeric values intensify or mitigate the probability of existence of a link. That implies treating triples with low-numeric values as ``pseudo-negatives''. Traditional embedding models put at the same level low-valued and high-valued triples. However, in many real-world scenarios it is important to make such distinction (e.g. predicting interactions in protein networks~\cite{PPI}).

We propose FocusE, an add-on layer for knowledge graph embeddings to enhance link prediction with edge-related numeric literals. FocusE works with any existing KGE model that adopts the standard negatives generation protocol~\cite{bordes2013translating}. 
We use edge numeric literals to modulate the margin between the scores of true triples and their corresponding negative corruptions.
Inspired by Focal Loss~\cite{Lin2017} that aims at sparser hard examples by modulating the loss function, we leverage numeric literals to ``focus'' traditional KGE models on triples with higher numeric values.

Experiments show that models trained using FocusE outperform numeric-unaware baselines, in particular in discriminating triples with high-numeric attributes from those associated to low values.

\section{Related Work}\label{sec:relatedwork}

\textbf{Knowledge Graph Embeddings.} Although a comprehensive survey is out of the scope of this work (recent surveys provide a good coverage of the landscape~\cite{bianchi2020knowledge}), it is worth listing the most popular knowledge graph embedding models proposed to date.
TransE~\cite{bordes2013translating} is the forerunner of distance-based models, and inspired a number of models commonly referred to as TransX.
The symmetric bilinear-diagonal model DistMult~\cite{yang2014embedding} paved the way for its asymmetric evolutions in the complex space, ComplEx~\cite{trouillon2016complex} and RotatE~\cite{sun2019rotate}.
Some models such as RESCAL~\cite{nickel2011three}, TuckER~\cite{balavzevic2019tucker}, and SimplE~\cite{kazemi_simplE} rely on different tensor decomposition techniques. 
Models such as ConvE~\cite{DBLP:conf/aaai/DettmersMS018} or ConvKB~\cite{nguyen2018novel} leverage convolutional layers. 
Attention is used by~\cite{nathani2019learning}. 

\textbf{Numeric-aware models (node attributes).}
None of the models listed above leverage numeric attributes of any kind. 
However, a number of recent works support multimodal knowledge graphs and learn from numeric values associated to node entities.
LiteralE enriches node embeddings with numeric information before scoring the triples~\cite{kristiadi2019incorporating}. KBLRN combines latent, relational and numeric features using product of experts model~\cite{garcia2017kblrn}. TransEA learns a vanilla structural model using TransE scoring, and an attribute model for attributed triples, using regression over the attribute values, which is jointly trained~\cite{wu2018knowledge}. 
Nevertheless, such models are not designed to learn from numeric values associated to \textit{edges}. 

\textbf{Numeric-aware models (edge attributes).}
To the best of our knowledge\footnote{We limit to Knowledge Graph Embeddings literature and knowledge graphs, i.e. directed, labeled, multi graphs. Other numeric-aware models such as graph neural networks that operate on different data modalities are out of the scope of this paper (they are either designed for tasks other than link prediction, most of them do not support multi-relational graphs, and cannot currently be applied at the scale KGE models operate). 
}, the only work designed to work with numeric-aware edges is UKGE~\cite{chen2019embedding}. 
UKGE generates confidence scores for known triples by squashing numeric values in the $[0-1]$ interval. 
It then uses probabilistic soft logic~\cite{kimmig2012short} to predict probability estimates for unseen triples, by jointly training a model to regress over the confidence values. A limitation of this approach is that out-of-band logical rules are required as additional input. %
It is also worth noting that UKGE design rationale aims at supporting uncertain knowledge graphs, i.e. graphs whose edge numeric values represent uncertainty. In this paper we aim at supporting generic numeric values, regardless of their semantics (uncertainty, strength, importance, etc).
Moreover, we claim that triples with high numeric values should have higher impact, and this should be explicitly modeled in the scoring or loss computation.

\begin{figure*}
\centering
\includegraphics[width=\textwidth]{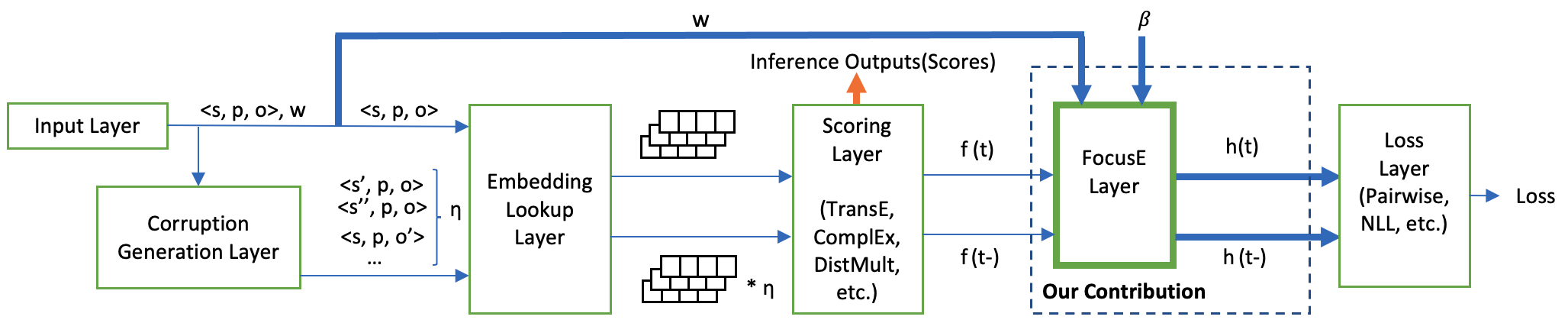}
\caption{A Knowledge Graph Embedding model architecture enhanced with FocusE. The add-on acts as an intermediate layer between the traditional scoring layer and the loss.}
\label{fig:arch}
\end{figure*}

\section{Preliminaries}\label{sec:background} 

\paragraph{Knowledge Graph.}
A knowledge graph $\mathcal{G}=\{ (s,p,o)\} \subseteq \mathcal{E} \times \mathcal{R} \times  \mathcal{E}$ is a set of triples $t=(s,p,o)$ each including a subject $s \in \mathcal{E}$, a predicate $p \in \mathcal{R}$, and an object $o \in \mathcal{E}$. $\mathcal{E}$ and $\mathcal{R}$ are the sets of all entities and relation types of $\mathcal{G}$.

\paragraph{Knowledge Graph with numeric-enriched triples.}
In a numeric-enriched knowledge graph $\mathcal{G}$, each triple is assigned a numeric attribute $w \in \mathbb{R}$, leading to $\mathcal{G}=\{ t=(s,p,o,w)\}$. It is worth noting that we are not tied to a specific numeric attribute semantics, as these numbers may encode importance, uncertainty, strength, etc. 
For example, Figure~\ref{fig:kg} assumes that numeric values indicate the importance of a link. The triple \texttt{(MLEngineer, requiresSkill, Python, 0.98)} is therefore more important than \texttt{(MLEngineer, requiresSkill, JavaScript, 0.13)}. 

Note $w$ can be either defined at a predicate level or at triple level. In this paper, we assume that $w$ is triple-specific. %

\paragraph{Knowledge Graph Embedding Models.}
Knowledge graph embedding models (KGE) are neural architectures designed to predict missing links between entities. KGE encode both entities $\mathcal{E}$ and relations $\mathcal{R}$ into low-dimensional, continuous vectors $\in \R^k$ (i.e, the embeddings). 
Knowledge graph embeddings are learned by training a neural architecture over a training knowledge graph: an input layer feeds training triples to an embedding lookup layer that retrieves embeddings of entities and relations. 

A scoring layer $f(t)$ assigns plausibility scores to each triple. The scoring layer is designed to assign high scores to positive triples and low scores to negative corruptions. 
Most of literature differs in the design rationale of $f(t)$.
For example, the scoring function of TransE~\cite{bordes2013translating} computes a similarity between the embedding of the subject $\mathbf{e}_{s}$ translated by the embedding of the predicate $\mathbf{e}_{p}$ and the embedding of the object $\mathbf{e}_{o}$. %

\textit{Corruptions} are synthetic negative triples generated by a corruption generation layer that follows the protocol proposed in~\cite{bordes2013translating}:
we define a corruption of $t$ as $t^-=(s,p,o')$ or $t^-=(s',p,o)$ where $s', o'$ are respectively subject or object corruptions (i.e. other entities randomly selected from $\mathcal{E}$). We generate synthetic negatives by corrupting one side of the triple at a time to comply with the local closed world assumption~\cite{nickel2016review}.

Finally, a loss layer optimizes the embeddings by maximizing the margin between positive triples $t$ and corruptions $t^-$. In other words, the goal of the optimization procedure is learning optimal embeddings, such that at inference time the scoring function $f(t)$ assigns high scores to triples likely to be correct and low scores to triples unlikely to be true.

\textbf{Link Prediction.} 
The task of predicting unseen triples in knowledge graphs is formalized in literature as a learning to rank problem, where the objective is learning a scoring function $f(t=(s, p, o)): \mathcal{E} \times \mathcal{R} \times \mathcal{E} \rightarrow \Real$ that given an input triple $t=(s,p,o)$ assigns a score $f(t) \in \Real$ proportional to the likelihood that the fact $t$ is true. 
Such predictions are ranked against predictions from synthetic corruptions, to gauge how well the model tells positives from negatives.

\paragraph{Link Prediction with numeric-enriched triples.}
In this paper we predict probability estimates for unseen numeric-enhanced triples $t=(s, p, o, w)$. 
The task is formalized as the same learning to rank problem described above for conventional link prediction.

\section{FocusE}\label{sec:method}

We present FocusE, an add-on layer for knowledge graph embedding architectures designed for link prediction with numeric-enriched triples.
FocusE takes into account numeric literals associated to each link. Regardless of their semantics, we operate under the assumption that numeric values intensify or mitigate the probability of existence of a link.
For example, given numeric values $w$ in the $[0-1]$ range, we assume that high values identify triples with higher chances of being true, low scores single out weak or unlikely relations, and $w=0$ triples are considered negative samples. 

FocusE consists in a plug-in layer that fits between the scoring and loss layers of a conventional KGE method and it is designed to be used during training (Figure~\ref{fig:arch}).
Unlike traditional architectures, before feeding the scoring layer to a loss function, we modulate its output based on numeric values associated to triples, to obtain ``focused'' scores. 
We leverage numeric values associated to triples so that during training the model focuses on triples with higher numeric values. 
We want our model to learn from training triples with high numeric values, and at the same time use edge numeric values to maximise the margin between scores assigned to true triples and those assigned to their corruptions. This increases the loss of the model and helps it focus on triples with higher values.
Our contribution is described below in detail.

Let $t=(s,p,o)$ be a positive triple and $w>0$ its numeric-value. We define a corruption of $t$ as $t^-=(s,p,o')$ or $t^-=(s',p,o)$. where $s', o'$ are respectively subject or object corruptions.

Let $f(t)$ be the scoring function of a KGE model. In the case of TransE~\cite{bordes2013translating} this is:

\begin{equation}
f(t) = -||\mathbf{e}_{s} + \mathbf{r}_{p} - \mathbf{e}_{o}||_n
\end{equation}

where $\mathbf{e}_{s}$, $\mathbf{r}_{p}$ and $\mathbf{e}_{o}$ are the embeddings of the subject $s$, predicate $p$, and object $o$.

We use a softplus non-linearity $\sigma$ to make sure the scores returned by $f(t)$ are greater or equal to zero, without introducing excessive distortion: %

\begin{equation}
g(t) = \sigma(f(t))= \ln{(1+e^{f(t)})} \geq 0
\end{equation}

To take into account the effect of numeric values associated to triples, we define a modulating factor $\alpha \in \mathbb{R}$ which is responsible to strike a balance between the influence of the structure of the graph and the impact of the numeric values associated to each triple:

\begin{equation}
\alpha =
\begin{cases}
 \beta + (1-w)(1-\beta) & \mbox{if \,} t \\
 \beta + w(1-\beta) & \mbox{if \,} t^-
\end{cases}
\label{eq:alpha}
\end{equation}

where $\beta \in [0,1]$ is the \textit{structural influence}, an hyperparameter that modulates the influence of graph topology, and $w \in \mathbb{R}$ is the numeric value associated to the \textit{positive} triple $t$. 
$\beta$ is used to re-weigh the triple value $w$. If $\beta=0$ the original numeric values $w$ are used. If $\beta=1$, numeric values $w$ are ignored and the model is equivalent to a conventional KGE architecture.
Note that positive and negative triples are assigned different $\alpha$ equations.
This is done to lower the margin between the scores for triples and their respective corruptions when the triple numeric value is high. 

Finally, the FocusE layer $h(t)$ is defined as:

\begin{equation}
h(t) = \alpha \, g(t)
\label{eq:ht}
\end{equation}

Putting all together, the FocusE layer $h(t)$ is then used in the loss function\footnote{Note that the FocusE layer described in Eq.\ref{eq:ht} is compatible with other losses used in KGE literature (e.g. pairwise, negative log-likelihood, etc.)} $L$. 
This is a modified, more numerically stable version of the negative log-likelihood of normalized softmax scores proposed in \cite{kadlec2017knowledge}:

\begin{equation}
L = - \sum_{t^+,t^-} log \frac{e^{h(t^+)}}{e^{h(t^+)} + e^{h(t^-)}} \label{eq:loss}
\end{equation}

As seen in \eqref{eq:loss}, the modulation between structural influence and numeric values increases the margin between high and low-valued triples. Hence the model learns to focus on triples with higher numeric attributes. 

During training, we decay the structural influence $\beta$ from 1 to 0. This is controlled by the hyperparameter $\lambda$ (decay). Initially the model gives equal importance to all training triples. When $\beta$ decays to zero (linearly, over $\lambda$ epochs), the model exclusively relies on numeric attributes ($w$).
Section~\ref{sec:decay} shows empirical evidence of the effectiveness of $\lambda$.

\section{Experiments}\label{sec:eval}
We assess the predictive power of FocusE on the link prediction task with numeric-enriched triples. Experiments show that FocusE outperforms conventional KGE models and its closest direct competitor UKGE~\cite{chen2019embedding} in discriminating low-valued triples from high-valued ones.

\textbf{Datasets}
We experiment with three publicly available benchmark datasets originally proposed by~\cite{chen2019embedding}. Triples are associated to numeric values $w$ interpreted as uncertainty values. We also introduce a fourth dataset, where numeric values $w$ encode the \textit{importance} of each link. 
All datasets include triple-specific $w$ values.
Table~\ref{table:kgstats} shows the statistics of all the datasets used for the experiments.

\begin{itemize}
  \item \textbf{CN15K}~\cite{chen2019embedding}. A subset of ConceptNet~\cite{CN}, a common sense knowledge graph built to represent general human knowledge. Numeric values on triples represent uncertainty.
  \item \textbf{NL27K}~\cite{chen2019embedding}. A subset of the Never-Ending Language Learning (NELL)~\cite{mitchell2018never} dataset, which collects data from web pages. Numeric values on triples represent link uncertainty.
  \item \textbf{PPI5K}~\cite{chen2019embedding}. Knowledge graph of protein-protein interactions~\cite{PPI}. Numeric values represent the confidence of the link based on existing scientific literature evidence.

  \item \textbf{O*NET20K}\footnote{\texttt{\url{https://docs.ampligraph.org/en/latest/ampligraph.datasets.html}}}. 
We introduce a subset of O*NET~\footnote{\texttt{\url{https://www.onetonline.org/}}}, a dataset that includes job descriptions, skills and labeled, binary relations between such concepts. Each triple is labeled with a numeric value that indicates the \textit{importance} of that link. 
Unlike the other datasets, we built a test set that includes a single predicate type that connects jobs to skills, under the assumption that we are interested in predicting that type of links only. This is done to mock up single-target link prediction tasks adopted in applied knowledge discovery scenarios.

\end{itemize} 

\begin{table}[t]
  \centering
  \footnotesize
  \setlength{\tabcolsep}{5pt}
  \begin{tabular}{l cccc}
      \toprule           & O*NET20K & CN15K & NL27K & PP15K\\ 
      \midrule
      Training              & 461,932 & 204,984 & 149,100 & 230,929            \\ 
      Validation$^\ddagger$  & 138 & 3532 & 8161 & 1940            \\ 
      \multirow{2}{*}{\shortstack[l]{Test \\(top 10\%)$^*$}}    & \multirow{2}{*}{200} & \multirow{2}{*}{1,929} & \multirow{2}{*}{1,402} & \multirow{2}{*}{2,172}       \\ 
      &&&& \\
      \multirow{2}{*}{\shortstack[l]{Test \\(bottom 10\%)$^*$}} & \multirow{2}{*}{200} & \multirow{2}{*}{1,929} & \multirow{2}{*}{1,402} & \multirow{2}{*}{2,172}       \\ 
      &&&& \\

      Entities              & 20,643 & 15,000 & 27,221 & 5,000 \\ 
      Relations             & 19 & 36 & 404 & 7 \\ 
      \bottomrule
  \end{tabular}
  \caption{Datasets used in experiments. ($^*$) test sets include either top-10\% valued triples or bottom 10\%. All $w$ are normalized, so that $w \in [0-1]$. ($^\ddagger$) validation sets only include high-valued triples where $w \geq 0.8$.}
  \label{table:kgstats}
\end{table}

\textbf{Implementation Details and Baselines.}
FocusE and all baselines are implemented with the AmpliGraph library~\cite{ampligraph} version 1.4.0, using TensorFlow 1.15.2 and Python 3.7. Code and experiments are available at \texttt{\url{https://github.com/Accenture/AmpliGraph}}.
We experiment with three popular KGE baselines: TransE, DistMult, ComplEx. For each baseline and for FocusE, we carried out extensive grid search, over the following ranges of hyperparameter values: embedding dimensionality $k=[200-600]$, with a step of 100; baseline losses=\{negative log-likelihood, multiclass-NLL, self-adversarial\}; synthetic negatives ratio $\eta=\{5, 10, 20, 30\}$; learning rate$=\{\num{1e-3}, \num{5e-3}, \num{1e-4}\}$; epochs$=[100-800]$, step of 100; L3 regularizer, with weight $\gamma=\{\num{1e-1}, \num{1e-2}, \num{1e-3}\}$.
For FocusE we also tuned the decay $\lambda=[100-800]$, with increments of 100. 
Best combinations for all experiments are reported in Appendix~\ref{app:hyper}.

For experiments with UKGE we modified\footnote{The original UKGE codebase does not perform a fair rank computation: the authors assign rank=1 when $t$ and $t^-$ have the same score, whereas we assign a lower rank, as per agreed upon practice in the community. For example, if we have $\eta=1,000$ corruptions $t^-$ which are all assigned the same score as $t$, we assign $t$ a rank=1,000 whereas UKGE ranks it first. To guarantee a fair comparison, we aligned UKGE to our procedure.} the original codebase\footnote{\texttt{\url{https://github.com/stasl0217/UKGE}}} provided by the authors to generate learning to rank metrics required by our experimental protocol. We used the best hyperparameter configuration proposed by the authors.
Note we do not evaluate UKGE on O*NET20K, since the model requires additional logical rules which are not available for this dataset.

All experiments were run under Ubuntu 16.04 on an Intel Xeon Gold 6142, 64 GB, equipped with a Tesla V100 16GB.

\textbf{Evaluation protocol.}
We adopt the standard evaluation protocol described by \cite{bordes2013translating} to all datasets described above. We predict whether each triple $t=(s,p,o) \in \mathcal{T}$ is a positive fact, where $\mathcal{T}$ is a disjoint held-out test set that includes only positives triples. 
We cast the problem as a learning-to-rank task: for each $t=(s,p,o) \in \mathcal{T}$, we generate synthetic negatives $t^- \in \mathcal{N}_t$ by corrupting one side of the triple at a time (i.e. either the subject or the object). 
We predict a score for each $t$ and all its negatives $t^- \in \mathcal{N}_t$. We then rank the only positive $t$ against all the negatives $\mathcal{N}_t$ \footnote{As for standard protocol, we used distinct entities twice, leading to CN15K=$\sim$30k, NL27K=$\sim$54k, PP15K=10k, O*NET20K=$\sim$40k synthetic negatives (Table~\ref{table:kgstats}).}. 
We report learning to rank metrics such as mean rank (MR), mean reciprocal rank (MRR), and Hits at $n$ (where $n=1, 10$) by filtering out spurious ground truth positives from the list of generated corruptions (i.e. ``filtered'' metrics).

\subsection{Predicting High-Valued Links}\label{sec.top10}
First, we assess how well FocusE predicts triples associated to \textit{high} numeric values. This is an important task in many applicative scenarios: high numeric values usually imply low uncertainty, high strength or importance, thus leading to more valuable newly-discovered knowledge.
Table~\ref{table:high-p} reports, for each dataset, the best mean rank, MRR, and Hits@\{1,10\} computed over test sets that include only triples associated to the top-10\% numeric attributes. Our goal is learning to assign low ranks to test triples (i.e. rank=1 being the best outcome).

Results show that FocusE brings better or very similar MRR to traditional, numeric-unaware baselines:
on O*NET20K, FocusE increases MRR for all models, and it outperforms the best baseline by 14 base points (MRR marginally differs on CN15K, NL27K, and PPI5K).

Experiments show that FocusE outperforms UKGE: on CN15K, MRR is 15 base points higher, 19 points higher on NL27K, and up to 30 points better for PPI15K (UKGE$_{rect}$ fails to provide actionable results\footnote{Due to a problem in the UKGE codebase, UKGE predictive power is MRR=0, hence "not actionable".} on PPI15K). 
It is worth mentioning that UKGE requires additional external rules, hence the absence of results for O*NET20K. FocusE achieves better predictive power, without requiring additional out-of-band rules.

\begin{table}
  \centering
  \footnotesize
  \setlength\tabcolsep{2pt}
  \begin{tabular}{ll cccc}

      \toprule           
      & & \multicolumn{4}{c}{\textbf{MRR} (bottom 10\%)} \\
      & & O*NET20K & CN15K & NL27K & PP15K\\ 
      \midrule
      TransE &         & .13 & .05 & .36 & .26            \\ 
      DistMult &       & .15 & .05 & .83 & .97             \\ 
      ComplEx&         & \textbf{.12} & .08 & .81 & .95       \\ 
      UKGE$_{rect}$ &  &  - & \textbf{.02} & .28 & .00$^*$ \\ 
      UKGE$_{logi}$ &  &  - & \textbf{.02} & .36 & .56 \\
      \multirow{3}{*}{\textbf{FocusE}}               
        & TransE       & .14  & .04 & \textbf{.26} & \textbf{.08}       \\ 
        & DistMult     & .15 & .05 & .70 & .72       \\ 
        & ComplEx      & .13 & .08 & .53 & .49       \\  
      \bottomrule
  \end{tabular}
  \caption{Predictive power of FocusE on the 10\% low-valued triples. Lower = better. Best results in bold. ($^*$) UKGE$_{rect}$ fails to produce actionable results on PPI5K.}
  \label{table:low-p}
\end{table}

\begin{table*}
  \centering
    \small

  \begin{tabular}{ll @{\extracolsep{8pt}} cccc  cccc}
      & & \multicolumn{3}{c}{\textbf{O*NET20K} (top 10\%)}  &  & & \multicolumn{3}{c}{\textbf{CN15K} (top 10\%)}       \\
      \cline{3-6} \cline{7-10}

      & & & & \multicolumn{2}{c}{Hits@}  &    & & \multicolumn{2}{c}{Hits@}    \\
      \cline{5-6} \cline{9-10}

      & & MR & MRR & 1  & 10 & MR & MRR & 1  & 10  \\

       \midrule

        \multicolumn{2}{l}{TransE} 
        & \underline{8}      & .37          & .00       & .88
        & \underline{623}     & .19          & .03       & \textbf{.47}  \\

        \multicolumn{2}{l}{DistMult}  
        & 13      & \underline{.49}         & \underline{.33}       & .82
        & 942      & .22      & .15       & .37     \\

        \multicolumn{2}{l}{ComplEx} 
        & 26      & .32          & .19      & .59
        & 863      & \underline{.28}      & \textbf{.22}       & .40     \\

        \multicolumn{2}{l}{UKGE$_{rect}$} 
        & -      & -          & -       & -
        & 984      & .14         & .05       & .32  \\

        \multicolumn{2}{l}{UKGE$_{logi}$} 
        & -      & -          & -       & -        
        & 2214      & .13         & .08       & .22  \\
        
        \midrule

        \multirow{3}{*}{\textbf{FocusE}}
        & TransE
        & \textbf{4}      & .41      & .0       & \textbf{.94}        
        & \textbf{438}      & .18      & .02     & \textbf{.47}     \\
        
        & DistMult
        & 102      & \textbf{.63}      & \textbf{.52}      & \underline{.85}         
        & 1552      & .24      & .17      & .37     \\
        
        & ComplEx
        & 9      & .46      & .23       & .83     
        & 840      & \textbf{.29}      & \underline{.21}       & \underline{.45}     \\

  \end{tabular}
  \caption*{}

  \begin{tabular}{ll @{\extracolsep{8pt}} cccc  cccc}
      & & \multicolumn{3}{c}{\textbf{NL27K} (top 10\%)}  &  & & \multicolumn{3}{c}{\textbf{PP15k} (top 10\%)}       \\
      \cline{3-6} \cline{7-10}

      & & & & \multicolumn{2}{c}{Hits@}    &  & & \multicolumn{2}{c}{Hits@}    \\
      \cline{5-6} \cline{9-10}

      & & MR & MRR & 1  & 10 & MR & MRR & 1  & 10  \\

       \midrule

        \multicolumn{2}{l}{TransE} 
        & \underline{91}      & .56          & .38       & .86         
        & 6      & .42          & .00       & .94     \\

        \multicolumn{2}{l}{DistMult}  
        & 130      & .83      & .77       & \underline{.93}        
        & 6      & \textbf{.97}      & \textbf{.96}       & \textbf{.99}     \\

        \multicolumn{2}{l}{ComplEx} 
        & \textbf{90}      & \textbf{.87}      & \textbf{.82}       & \textbf{.94}        
        & 10      & \textbf{.97}      & \underline{.95}      & .\textbf{.99}     \\
        
        \multicolumn{2}{l}{UKGE$_{rect}$} 

        & 169      & .61      & .50      & .80     
        & 4943    & .00        & .00         & .00 \\

        \multicolumn{2}{l}{UKGE$_{logi}$} 
        & 299      & .65         & .56       & .80
        & \textbf{2}      & .65        & .37      & \textbf{.99}  \\

        \midrule

        \multirow{3}{*}{\textbf{FocusE}}
        & TransE
        & \textbf{90}     & .57      & .37       & .88         
        & \underline{4}      & .43      & $<.01$       & \underline{.96}     \\
        
        & DistMult
        & 140      & .82      & .77       & .92         
        & 16      & \underline{.96}      & .93       & \textbf{.99}     \\
        
        & ComplEx
        & 224     & \underline{.84}      & \underline{.79}     & .92         
        & 20      & .95      & .92       & \textbf{.99}     \\

  \end{tabular}

\caption{Predicting high-valued links: ranking metrics computed on test sets that consist in the top-valued 10\% triples. Filtered metrics. Best results in bold, second best underlined.}
\label{table:high-p}
\end{table*}

\begin{table*}
  \centering
    \small

  \begin{tabular}{ll @{\extracolsep{8pt}} ccc  ccc}
      & & \multicolumn{3}{c}{\textbf{O*NET20K}} & \multicolumn{3}{c}{\textbf{CN15K}}       \\
      \cline{3-5} \cline{6-8}

      & & \multirow{2}{*}{\shortstack[c]{MRR\\(top 10\%)}} & \multirow{2}{*}{\shortstack[c]{MRR\\(bottom 10\%)}}   & \multirow{2}{*}{$\Delta$MRR} 
      & \multirow{2}{*}{\shortstack[c]{MRR\\(top 10\%)}} & \multirow{2}{*}{\shortstack[c]{MRR\\(bottom 10\%)}}   & \multirow{2}{*}{$\Delta$MRR}\\ 
      &&&& \\
\midrule
        \multicolumn{2}{l}{TransE} 
        & .37      & .13          & .24  
        & .19      & .05          & \textbf{.14}   \\

        \multicolumn{2}{l}{\textbf{FocusE} TransE} 
        & .41      & .14          & \textbf{.27 (+13\%)} 
        & .18      & .04          & \textbf{.14}   \\
        
        \midrule

        \multicolumn{2}{l}{DistMult}  
        & .49      & .15          & .34  
        & .22      & .05          & .17   \\
        
        \multicolumn{2}{l}{\textbf{FocusE} DistMult} 
        & .63      & .15          & \textbf{.48 (+41\%)} 
        & .24      & .05          & \textbf{.19 (+12\%)}   \\
        
        \midrule

        \multicolumn{2}{l}{ComplEx} 
        & .32      & .12          & .20  
        & .28      & .08         & .20   \\
        
         \multicolumn{2}{l}{\textbf{FocusE} ComplEx} 
        & .46      & .13         & \textbf{.33 (+65\%)} 
        & .29      & .08          & \textbf{.21 (+5\%)}   \\

        \midrule

        \multicolumn{2}{l}{UKGE$_{rect}$} 
        & -      & -         & - 
        & .14      & .02         & .12   \\

        \multicolumn{2}{l}{UKGE$_{logi}$} 
        & -      & -         & - 
        & .13      & .02         & .11   \\
        
         \multicolumn{2}{l}{\textbf{FocusE} ComplEx} 
        & .46      & .13         & .33 
        & .29      & .08          & \textbf{.21 (+75\%)}   \\

  \end{tabular}
    \caption*{}
  
  \begin{tabular}{ll @{\extracolsep{8pt}} ccc  ccc}
      & & \multicolumn{3}{c}{\textbf{NL27K}} & \multicolumn{3}{c}{\textbf{PPI5K}}       \\
      \cline{3-5} \cline{6-8}

      & & \multirow{2}{*}{\shortstack[c]{MRR\\(top 10\%)}} & \multirow{2}{*}{\shortstack[c]{MRR\\(bottom 10\%)}}   & \multirow{2}{*}{$\Delta$MRR} 
      & \multirow{2}{*}{\shortstack[c]{MRR\\(top 10\%)}} & \multirow{2}{*}{\shortstack[c]{MRR\\(bottom 10\%)}}   & \multirow{2}{*}{$\Delta$MRR}\\ 
      &&&& \\
\midrule
        \multicolumn{2}{l}{TransE} 
        & .56      & .36          & .20  
        & .42      & .26         & .16   \\
        
        \multicolumn{2}{l}{\textbf{FocusE} TransE} 
        & .57      & .26          & \textbf{.31 (+55\%)}
        & .43      & .08          & \textbf{.35 (+119\%)}   \\
        
        \midrule
        
        \multicolumn{2}{l}{DistMult}  
        & .83      & .83          & $<.01$  
        & .97      & .97         & $<.01$   \\
       
        \multicolumn{2}{l}{\textbf{FocusE} DistMult}  
        & .82      & .70          & \textbf{.12 (+1,100\%)}  
        & .96      & .72          & \textbf{.24 (+2,300\%)}   \\
        
        \midrule
        
        \multicolumn{2}{l}{ComplEx} 
        & .87      & .81          & .06  
        & .97      & .95          & .02   \\
        
        \multicolumn{2}{l}{\textbf{FocusE} ComplEx} 
        & .84      & .53          & \textbf{.31 (+417\%)}  
        & .95      & .49          & \textbf{.46 (+2,200\%)}   \\

         \midrule

        \multicolumn{2}{l}{UKGE$_{rect}$} 
        & .61      & .28         & \textbf{.33} 
        & .00      & .00         & .00   \\

        \multicolumn{2}{l}{UKGE$_{logi}$} 
        & .65      & .36         & .29 
        & .65      & .56         & .09   \\
        
         \multicolumn{2}{l}{\textbf{FocusE} ComplEx} 
        & .84      & .53          & .31 (-6\%)  
        & .95      & .49          & \textbf{.46 (+411\%)}   \\
          
  \end{tabular}

\caption{High-valued vs low-valued triples discriminative power: FocusE brings larger differences in MRR across the board, showing better capabilities at correctly ranking the top-10\% high-valued triples vs the bottom 10\%. Best results in bold.}

\label{table:deltaMRR}
\end{table*}

\subsection{Discriminating High and Low-Valued Links}
The primary goal of FocusE is achieving a clear separation between scores assigned to high and low-valued triples. %
To assess how well FocusE can tell high-valued links from low-valued, we look at two complementary tasks: how the model performs on test triples with high values (top 10\%) and on triples with low values (bottom 10\%). We collect the respective MRR results and we compute the difference $\Delta$MRR. The higher the difference, the better. 
Note this task differs from what traditionally done by KGE models, which are designed to tell positives from synthetic negatives only, thus not being able to discriminate between low and high-valued triples.

Such prior art shortcoming is evident from results in Table~\ref{table:deltaMRR}. Models trained using FocusE have a much wider $\Delta$MRR compared to traditional baselines. 
Using link numeric values of training triples, FocusE better discriminates triples with higher values in the test set from the ones with lower values (see also Table~\ref{table:low-p}  for details on the bottom 10\% links). 
In particular, on O*NET20K, NL27K and PPI5K, FocusE models perform exceptionally better, with $\Delta$MRR being consistently higher than traditional KGE baselines. 
FocusE also outperforms UKGE by a large margin (75\%) on CN15K. It provides a large margin of 40 base points on PPI5K as well, i.e. a 411\% increase (UKGE$_{rect}$ fails to provide actionable results on PPI15K).

Appendix \ref{app:violin} shows additional evidence of FocusE outperforming baselines, by comparing distributions of scores (logits).

\subsection{Decay Impact}\label{sec:decay}

The structural influence $\beta$ affects the impact of numeric values $w$. If $\beta=1$, FocusE falls back to the underlying KGE model. If $\beta=0$, the original numeric values $w$ have maximum impact throughout the entire training. 
During training we decay $\beta$ linearly, until it reaches 0. 

\begin{figure}[t]
\centering
\includegraphics[width=.7\columnwidth]{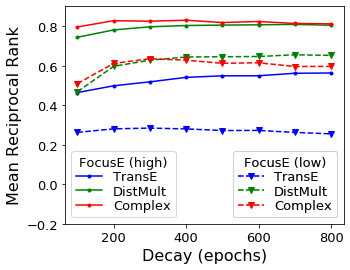}
\caption{Impact of $\lambda$, the decay of structural influence $\beta$ on model performance over n-epochs for NL27k dataset.}
\label{fig:decay}
\end{figure}

In this experiment we vary the decay $\lambda$, while freezing the other hyperparameters to ($k = 300$, $\eta=30$, NLL of normalized softmax scores loss (Equation~\ref{eq:loss}), Adam optimizer with learning rate \num{1e-3}, L3 regularizer, with weight $\gamma=\num{1e-2}$. 
We choose different decay epoch values $\lambda=\{0, 100, 200, 300, 400, 500, 600, 700, 800\}$.
We report changes in MRR on NL27K's top 10\% test set by experimenting on TransE, DistMult, and ComplEx. 
Figure~\ref{fig:decay} shows that performance improves if $\lambda$ increases. In most cases model performance saturates when $\lambda > 400$ epochs. %

\section{Conclusion}\label{sec:conclusion}
We show that by plugging an additional layer we can make a conventional KGE architecture aware of numeric values associated to triples. This leads to models that better discriminate high-valued and low-valued triples, regardless of the semantics of the numeric attributes, and without requiring additional out-of-band rules (unlike UKGE).

Future work will investigate the capability of predicting numeric values associated to unseen triples. We will also extend our approach to support multiple numeric attributes associated to the same triple.

\bibliographystyle{named}
\bibliography{references}

\clearpage
\onecolumn

\section*{Supplementary Material: Learning Embeddings from Knowledge Graphs With Numeric Edge Attributes}
\appendix

\section{Discriminating High and Low Valued Links: Scores Distribution}\label{app:violin}

\begin{figure}[H]
    \centering
    \begin{subfigure}[b]{0.35\textwidth}
        \includegraphics[width=\textwidth]{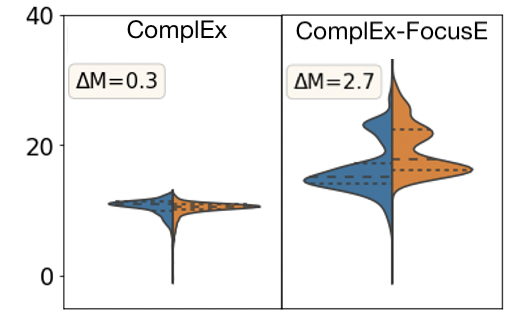}
        \caption{PP15K}
        \label{fig:ppi5kviolin}
    \end{subfigure}
    ~ %
    \begin{subfigure}[b]{0.35\textwidth}
        \includegraphics[width=\textwidth]{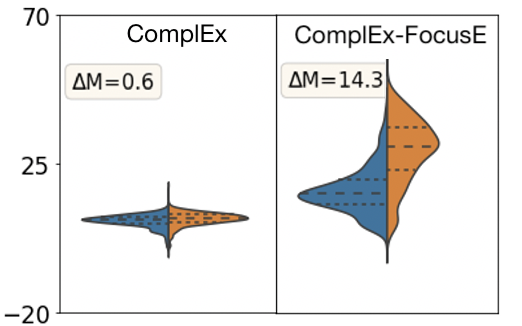}
        \caption{NL27K}
        \label{fig:nl27kviolin}
    \end{subfigure}
    ~ %
    \begin{subfigure}[b]{0.35\textwidth}
        \includegraphics[width=\textwidth]{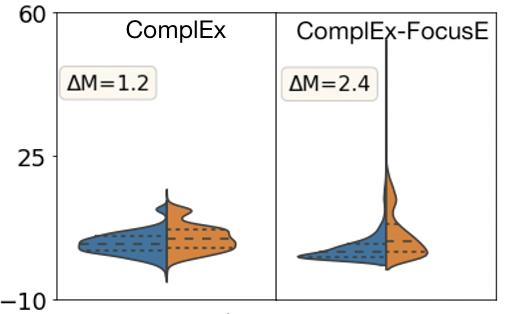}
        \caption{CN15K}
        \label{fig:cn15kviolin}
    \end{subfigure}
    \begin{subfigure}[b]{0.35\textwidth}
        \includegraphics[width=\textwidth]{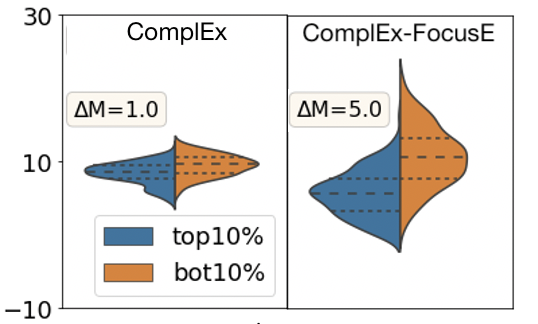}
        \caption{O*NET20K}
        \label{fig:onet20violin}
    \end{subfigure}
    \caption{Comparison of scores for plain ComplEx (left) and its FocusE counterpart (right), for top 10\% (orange) and bottom 10\% (blue) triples for different datasets. Dotted lines in score distributions indicate the 3 quartiles. $\Delta M$ is the absolute difference of the median of top 10\% and bottom 10\% scores.}\label{fig:violins}
\end{figure}

Figure~\ref{fig:violins} shows the distribution of the scores (logits) returned by a KGE model, for the top 10\% (orange) and bottom 10\% (blue) test set triples. The violin plots for the ComplEx models and their FocusE counterparts also report $\Delta M$ values, i.e. the absolute difference between the median of top 10\% and bottom 10\% scores. The higher this value, the better the model is at discriminating low from high-valued links. We see that for all datasets FocusE outperforms the baseline. Hence, models trained with FocusE are better at discriminating between triples with high numeric values (importance / uncertainty / etc.) from those with low values.

\section{Hyper-parameters}\label{app:hyper}
Table \ref{table:hyperparams} includes the hyperparameter values of the models reported in tables \ref{table:high-p} and \ref{table:deltaMRR}. For all experiments we used Adam with L3 regularizer (weight $\gamma$). The number of synthetic negatives per positive used in training is determined by $\eta$.

\begin{table}[H]
  \centering
  \scriptsize
  \setlength{\tabcolsep}{5pt}
  \begin{tabular}{l ccccc | ccccc}
      \toprule      

      & & \multicolumn{4}{c}{\textbf{Plain}} & \multicolumn{5}{c}{\textbf{w/ FocusE}} \\
      \cline{3-6} \cline{7-11}

      &  & k &Learning Rate &$\eta$ & $\gamma$ & k &Learning Rate &$\eta$ & $\gamma$ & $\lambda$ (Decay of $\beta$ [epochs])\\ 
      \midrule

      \multirow{3}{*}{\textbf{O*NET20K} }
          &TransE &400 &1e-3 &30 &1e-3 &400 &1e-3 &30 &1e-3 &800\\
          &DistMult &400 &1e-3 &30 &1e-3 &400 &1e-3 &30 &1e-3 &800 \\
          &ComplEx &400 &5e-5 &30 &1e-3 &400 &1e-3 &30 &1e-3 &500 \\
      \midrule
      \multirow{3}{*}{\textbf{CN15K} } 
          &TransE &400 &1e-5 &30 &1e-3 &250 &1e-3 &30 &1e-2 &800 \\
          &DistMult &400 &1e-4 &30 &1e-3 &400 &1e-4 &30 &1e-3 &800 \\
          &ComplEx &400 &1e-4 &30 &1e-3 &800 &1e-3 &30 &1e-2 &300 \\
      \midrule    
      \multirow{3}{*}{\textbf{NL27K} } 
          &TransE &200 &5e-5 &30 &1e-3 &200 &1e-3 &30 &1e-2 &800 \\
          &DistMult &400 &1e-4 &30 &1e-3  &500 &1e-3 &30 &1e-3 &200\\
          &ComplEx &400 &1e-4 &30 &1e-3 &400 &1e-4 &30 &1e-3 &800 \\
      \midrule    
      \multirow{3}{*}{\textbf{PPI5k} }
          &TransE &600 &1e-4 &30 &1e-2 &150 &1e-3 &30 &1e-1 &800 \\
          &DistMult &400 &1e-4 &30 &1e-2 &400 &5e-4 &30 &1e-3 &200 \\
          &ComplEx &400 &1e-4 &30 &1e-2 &400 &1e-4 &30 &1e-3 &500 \\
          
      \bottomrule
  \end{tabular}
  \caption{Hyper-parameters of the baseline KGE models and their FocusE counterparts of experiments reported in Section\ref{sec:eval}.} 
  \label{table:hyperparams}
\end{table}

\end{document}